**The Design and Evaluation of an Obstacle Avoidance Algorithm in Autonomous Vehicles with Drivers' Subjective Feelings Considered**

**Weishun Deng**
Department of Civil and Environmental Engineering
The Hong Kong University of Science and Technology, Hong Kong SAR, China, 999077
Email: wdengam@connect.ust.hk

**Fan Yu**
School of Mechanical Engineering
Shanghai Jiao Tong University, Shanghai, China, 200240
Email: fanyu@sjtu.edu.cn

**Zhe Wang**
Department of Civil and Environmental Engineering
The Hong Kong University of Science and Technology, Hong Kong SAR, China, 999077
Email: cezhewang@ust.hk

**Dengbo He (Corresponding Author)**
Intelligent Transportation Thrust, Systems Hub
Robotics and Autonomous Systems Thrust, Systems Hub
The Hong Kong University of Science and Technology (Guangzhou), Guangdong, China, 511400
Department of Civil and Environmental Engineering
The Hong Kong University of Science and Technology, Hong Kong SAR, China, 999077
Email: dengbohe@ust.hk

Word Count: 3502 words + 3 table (250 words per table) = 4,252 words

*Submitted [Submission Date]*

**ABSTRACT**

With the fast development of driving automation technologies, users' psychological acceptance of driving automation has become one of the major obstacles to the adoption of the driving automation technology. The most basic function of a passenger car is to transport passengers or drivers to their destinations safely and comfortably. Thus, the design of the driving automation should not just guarantee the safety of vehicle operation but also ensure occupants' subjective level of comfort. Hence this paper proposes a local path planning algorithm for obstacle avoidance with occupants' subjective feelings considered. Firstly, turning and obstacle avoidance conditions are designed, and four classifiers in machine learning are used to respectively establish subjective and objective evaluation models that link the objective vehicle dynamics parameters and occupants' subjective confidence. Then, two potential fields are established based on the artificial potential field, reflecting the psychological feeling of drivers on obstacles and road boundaries. Accordingly, a path planning algorithm and a path tracking algorithm are designed respectively based on model predictive control, and the psychological safety boundary and the optimal classifier are used as part of cost functions. Finally, co-simulations of MATLAB/Simulink and CarSim are carried out. The results confirm the effectiveness of the proposed control algorithm, which can avoid obstacles satisfactorily and improve the psychological feeling of occupants effectively.

**Keywords:** Autonomous Vehicles, Human Factors, Model Predictive Control, Local Path Planning and Tracking

**INTRODUCTION**

Motion control is a vital part of driving automation. According to *(1)*, psychological instead of technological concerns might be a major obstacle to the comertialization of autonomous vehicles (AVs). Customers' acceptance of AVs and others' tolerance of the technique willcan be influenced by their level of trust in the technology. According to a survey by Shariff et al *(2)*, due to the lack of trust in the capability of the autonomous driving systems, seventy-eight percent of Americans are afraid to take ride in an AV, and only nineteen percent are willing to have a try. Thus, AV should guarantee not just objective safety of the vehicle, but also subjective feeling of safety in challenging urban road environment.

To understand how vehicle motions and control algorithms would affect users' acceptance of the AVs, a field test was conducted by Abraham et al. *(3)*. In this study, 300 students took a ride in Level 3 AVs and experienced nine typical driving scenarios. Participants' trust toward the AV, perceived availability of the AV, perceived ease of use, perceived safety, behavior intention, and willing to ride the AV again before and after taking the AV rides were recorded. Results show that these factors have great influence on the acceptance of Avs *(4)*. A prediction model for the acceptance of AVs was developed after an analysis of the elements influencing the acceptance of AVs.

To increase drivers' acceptance of AV, Schwarting et al *(5)* proposed that autonomous vehicles should comprehend the intentions of human drivers and adapt to their driving styles, incorporate social psychology into the decision-making process, and predict other road users' behaviors in order to plan the path of AV. For example, AVs should take the social value orientation into consideration, and quantify the degree of egoism and altruism when interacting with other road agents. In aother study by Keen st al. *(6)*, researchers found that humanizing the execution of the AVs (i.e., making the paths planned by AVs as close to the paths chosen by human drivers as possible) would improve the comfort of the occupants. Noriyasu Noto makes use the actual obstacle avoidance test path data of three drivers *(7)*. By modifying the potential field parameters of the artificial potential field, the paths planned based on the potential field have the maneuvering characteristics of the driver, and the driver's characteristics are considered in the vehicle motion control algorithm *(8)*. Bansal's end-to-end machine learning approach enables the system to imitate the driving patterns of human drivers. In fact, this type of approach is a further step in meeting the driver's personalized needs for autonomous vehicles *(9)*.

Further, the control algorithms of AVs should reflect driver's social preferences by modeling the driver's personality, and risk tolerance into the decision module. Thus, in this study, we used the driver's trust in AVs as a premise and proposed an evaluation index, the confidence index, designed to depict drivers' safety feeling of AVs. For instance, when a driver is not confcent in a control of an AV, the driver will make unnecessary or inappropriate intervention, even if the manuveuor by AV is correct.

In order to take the driver's trust and riding feeling into consideration, a novel local path planning and tracking algorithm is proposed. We obtained objective vehicle dynamics parameters and driver's subjective feelings through two typical lateral motion tests, turning and obstacle avoidance. Then established a subjective and objective evaluation model between the two using four classifiers, and the artificial potential field was used to establish a model representing passengers' psychological feelings towards obstacles and roads. Afterwards, we used the subjective and objective evaluation model and the two potential field functions as the cost functions of the MPC-based path planner. The constraints of vehicle state variables and the driver's psychological last point to steer are also combined as the constraints of the MPC-based path planner. Finally, an obstacle avoidance path is generated by path planner and passed to the path tracker, which outputs the front wheel turning angle to the controlled vehicle to control the vehicle to complete obstacle avoidance. In the whole process, the driver's ride feeling and the trust of the automatic driving are fully considered. The overall architecture of MPC-based local obstacle-avoidance controller is shown in Figure 1.

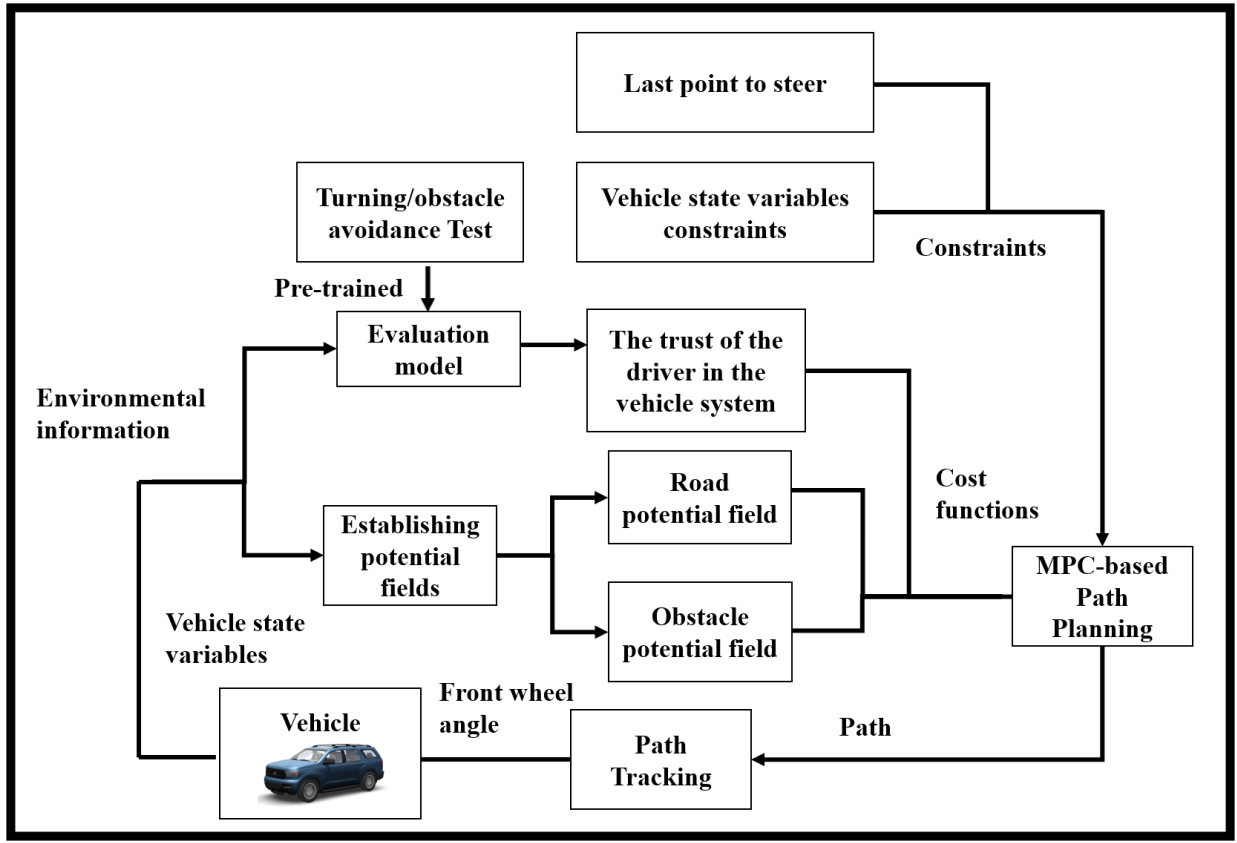

**Figure 1 The overall architecture of the MPC controller**

## METHODS

Urban roadways' typical turning and obstacle avoidance situations are designed to test the viability of the confidence index. Four classifiers are used to create an evaluation model that incorporates both the driver subjective confidence in the real track test and the objective vehicle system dynamics parameters.

### Field test

As seen in Figure 2 and 3, the typical obstacle avoidance and turning situations found on urban roadways are chosen as the test scheme to achieve more representative test results. In the turning conditions, the driver is required to traverse through the curve at the highest speed at which the driver believes they can safely pass the curve, as indicated in Figure 2. In the obstacle avoidance circumstances, the driver is required to approach the obstacle as soon as possible and avoid the obstacle only by steering, as seen in Figure 3. The purpose of these two tests is to provide a discernable subjective evaluation of the driver under the constraints of psychological limitations.

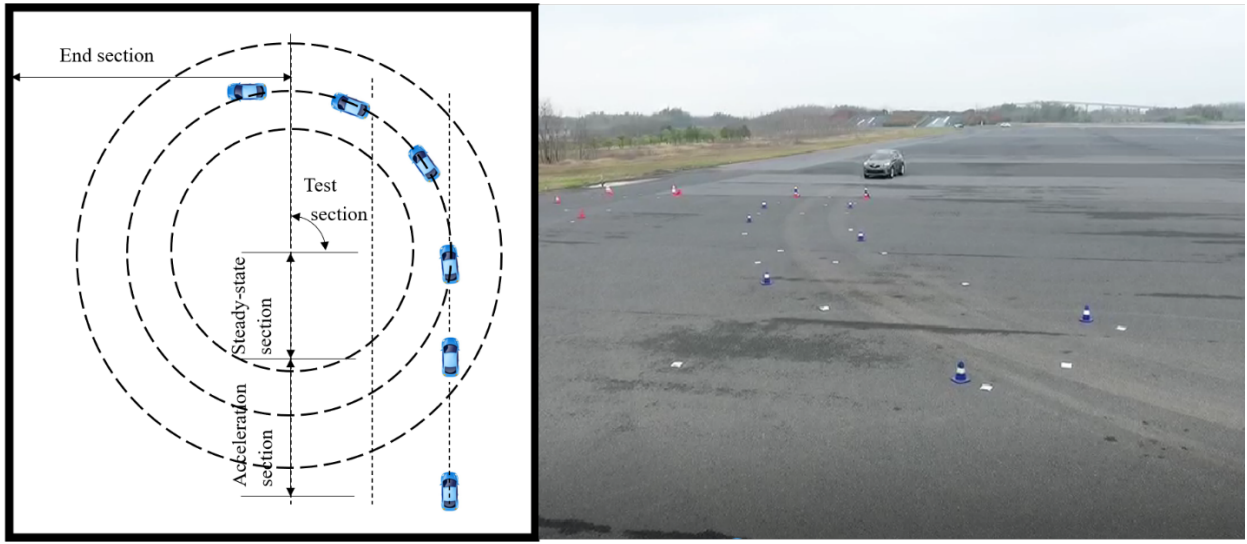

**Figure 2 Turning test**

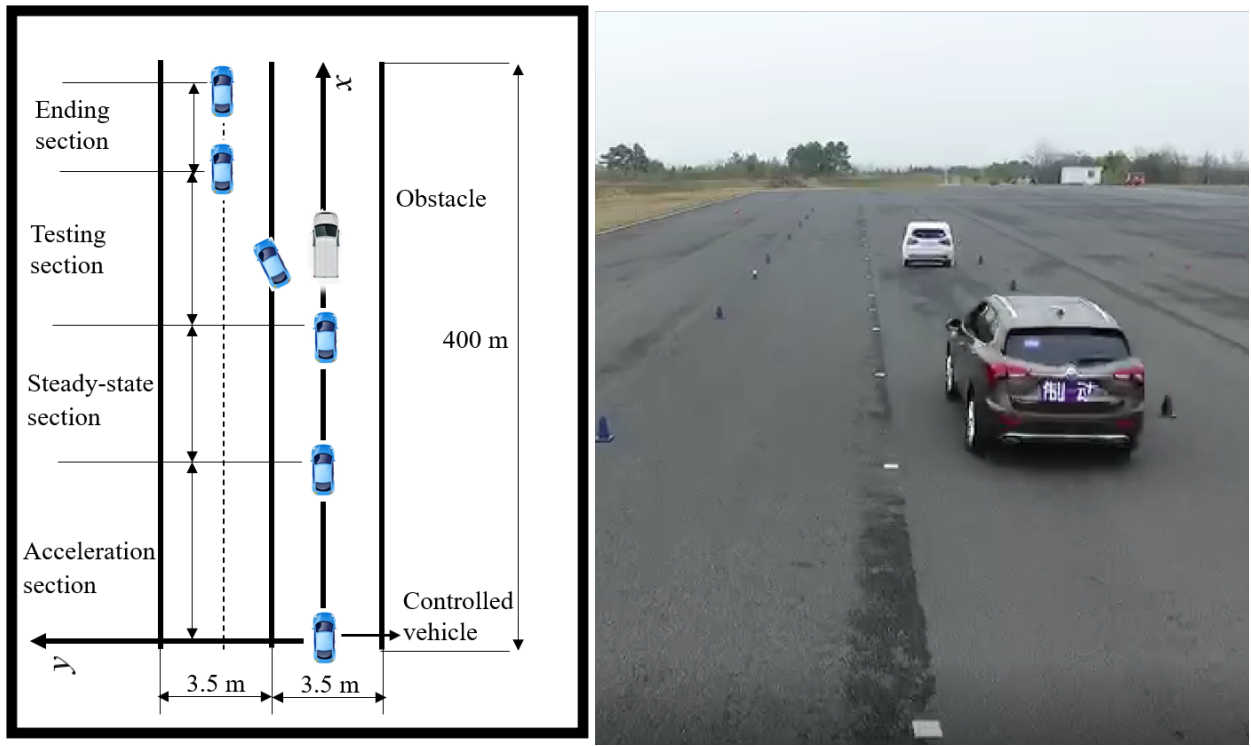

**Figure 3 Obstacle avoidance test**

The evaluation model establishes the relationship between objective dynamics parameters and subjective evaluation. Meanwhile, the evaluation model is established by four classifiers. Because the classifiers are continuous functions, the subsequent MPC algorithm can be used in conjunction with the evaluation model.

In terms of the psychological feelings of drivers and passengers, the influence factor can be divided into three aspects: vehicles, roads, and human. Hardware performance components like suspension, tires, and seats make up the majority of the vehicle-related elements. The roughness of the road surface is the main source of road factors. The driver's operating instructions are primarily responsible for the human factor in the driving environment. Whereas the driver's control inputs typically affect the longitudinal and lateral directions, the suspension and road roughness predominantly affect the vertical direction.

As a result, the integration of human factors with autonomous driving is crucial for the design and advancement of this technology. As shown in Figure 4, the three components of human factors in autonomous vehicles are: (1) whether the vehicle system control commands are consistent with the driver's driving habits; (2) the passenger's psychological feeling of obstacles and roads; and (3) the passenger's trust in autonomous vehicles.

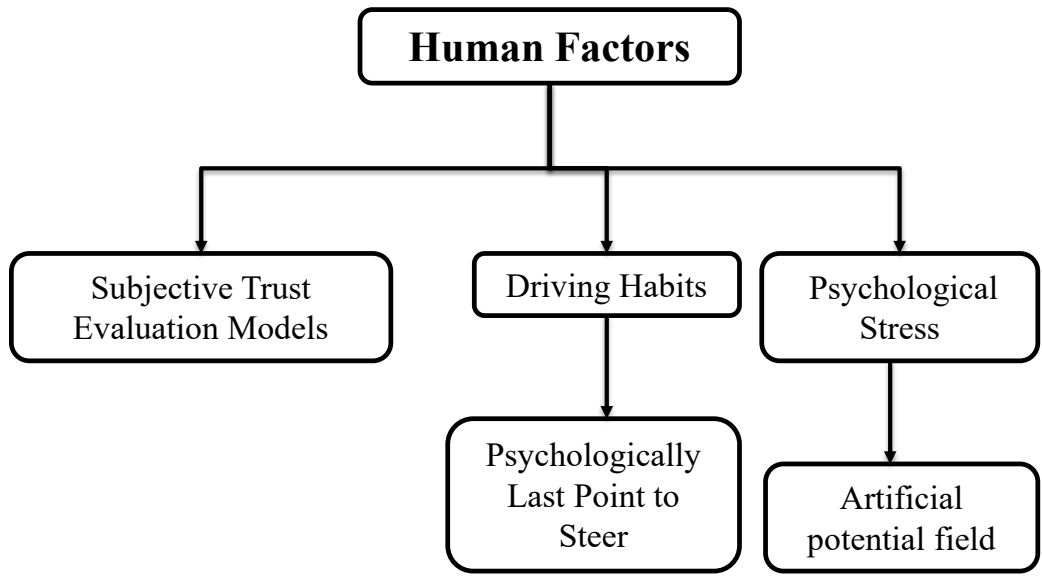

**Figure 4 Human factors**

The evaluation model is divided into two components: objective dynamic parameters serve as the input, and the corresponding output is subjective questionnaire scores. Then the model is trained according to the input and output/

**Subjective questionnaire**

Five items make up the subjective evaluation questionnaire, as seen in Figure 5. The weights of the five questions counterclockwise, as determined by the analytic hierarchy process (AHP) *(10)*, are 0.09, 0.18, 0.18, 0.18, and 0.37, respectively. Finally, the subjective evaluation's overall score is determined.

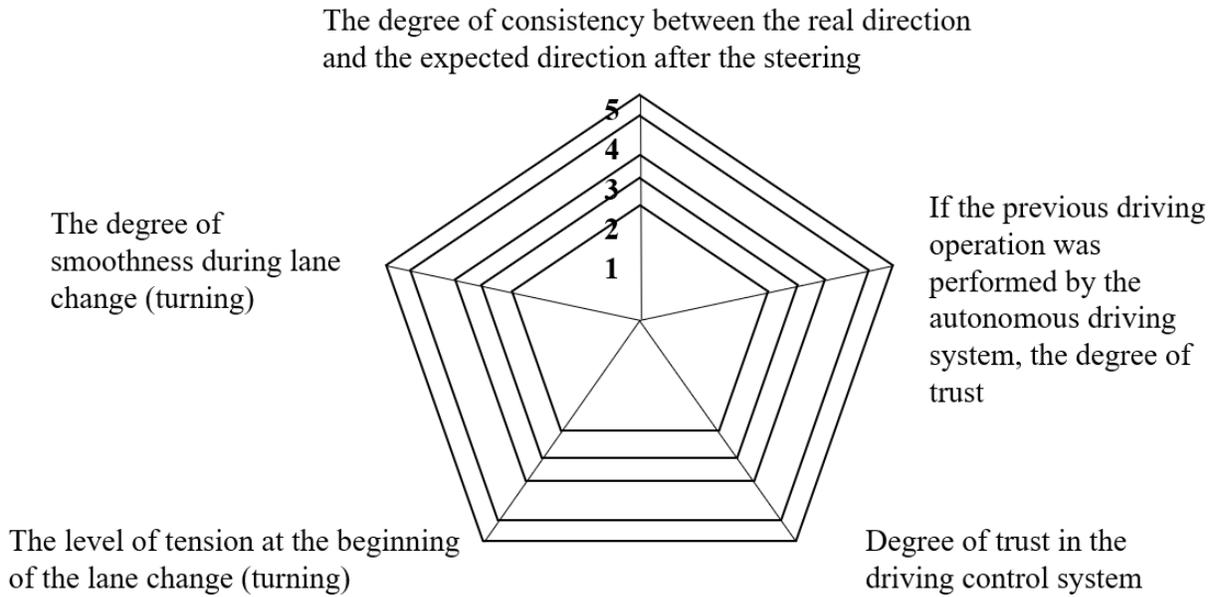

**Figure 5 Subjective evaluation questionnaire**

**Objective dynamics parameters**

Based on the correlation analysis, objective indicators that are more relevant to subjective evaluation are chosen. Table 1 displays the screening findings.

**TABLE 1 Objective Indices**

| Numbers | Indices |
|---------|---------|
| 1 | Longitudinal speed |
| 2 | Lateral acceleration |
| 3 | Yaw rate |
| 4 | Changing rate of lateral acceleration |
| 5 | Yaw angular acceleration |

Driver scores are divided into three kinds: good, normal, and poor, based on the results of the drivers' subjective questionnaire. Meanwhile, machine learning's classifier is employed *(11)*. Table 2 displays the accuracy of the four classifiers.

**TABLE 2 Objective Indices**

| Classifiers | Accuracy |
|-------------|----------|
| Template matching | 66.67% |
| Euclidean distance based on the class center | 62.12% |
| Mahalanobis distance | 68.18% |
| A Bayesian classifier based on minimum risk | 57.58% |

**Evaluation model**

　　Due to the high rental and personnel costs of the testing ground as well as the limited number of test data, this research employs the classifier in machine learning to establish the subjective and objective evaluation model.

　　The sample data used in the classifier is used to train a model. However, this training differs from the training of a neural network in that the latter employs self-feedback to fix internal parameters and reduce classification error. The classifier must continuously adjust the discriminant function artificially in order to distinguish the point sets of different classes and create a linear or nonlinear discriminant function.

　　Finally, the error of the classification result of the training samples can be minimized. The schematic diagram of the structure of the classifier based on the discriminant function is shown in Figure 6.

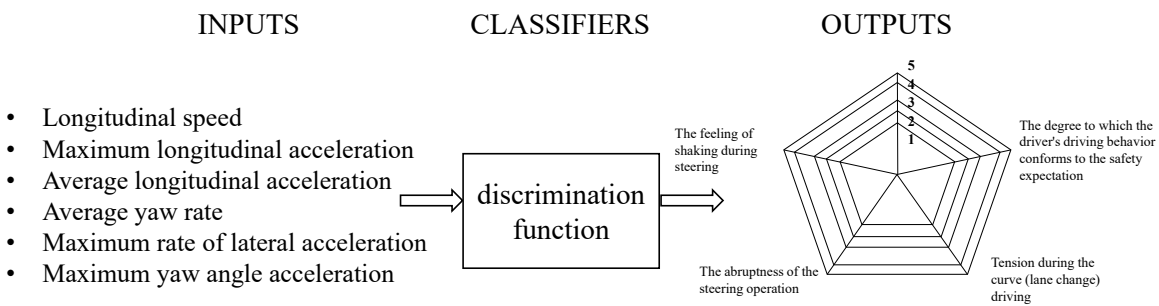

**Figure 6 Structure diagram of classifiers**

**Last point to steer**

　　The scene of the control algorithm is the same as the last test part, which is the obstacle avoidance scene, as shown in Figure 7. According to the physical meaning, a physically last point to steer (LPTS) can be obtained by the kinematic calculation. However, drivers will also have a matching psychological last point to steer (PLPTS) based on drivers' own experience and assessment for their psychological expectations, as shown in Figure 7. Therefore, for the control algorithm of autonomous vehicles, these two last points to steer should be satisfied at the same time.

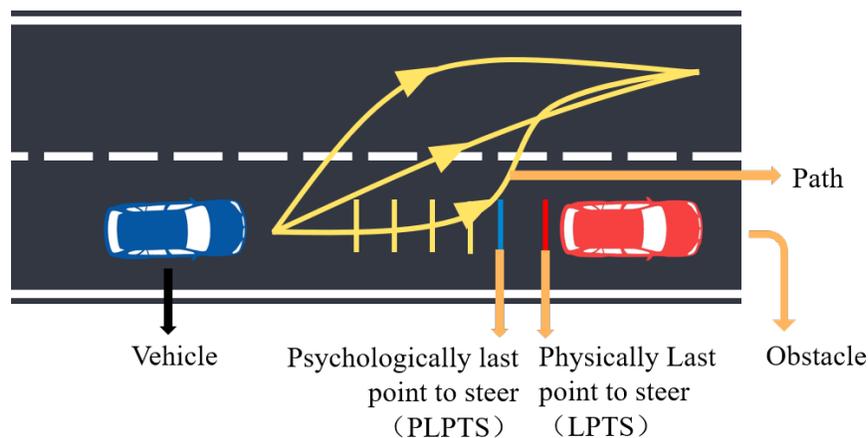

**Figure 7 Obstacle avoidance control algorithm**

**Artificial potential field**

When the speed was 10, 30, 60, and 80 km/h, the distance between PLPTS and the obstacle was 7.67, 8.65, 15.27, and 18.15 m respectively *(12)*. These distances will serve as constraints of the control algorithm. The path planning algorithm used by autonomous vehicles is inspired by robots, but because it is a manned vehicle, it must contend with more complex road circumstances than do robots. Therefore, the driving experience of the drivers and passengers must be taken into account when controlling autonomous vehicles. The artificial potential field is applied to create a numerical model because the potential field value is related to the relative speed and position of the vehicle and the obstacle, and the psychological pressure of the drivers and passengers on the obstacle and the road boundary is also related to the relative speed and position *(13)*. Figure 8 depicts the psychological pressure reactions of drivers and passengers to obstacles and road boundaries. Yellow means the function value is large, while blue means the function value is small.

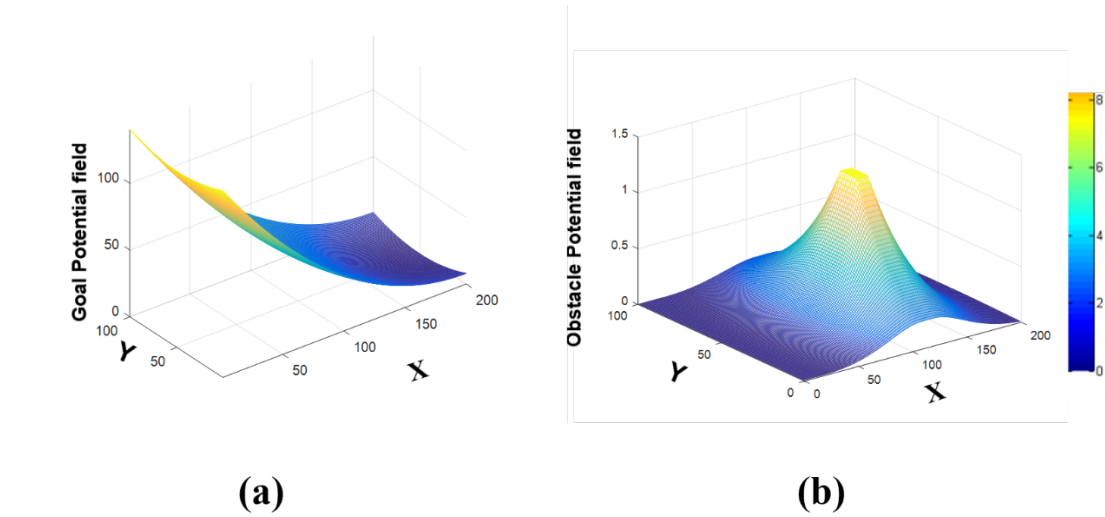

**(a)**                          **(b)**

**Figure 8 Artificial potential fields of (a) target points and (b) obstacles**

The driver's future control commands are based on the current environment, which is the same as the controlling idea of MPC. The above-mentioned driver confidence in autonomous vehicles is taken into account by the control algorithm.

We aim to consider the feasibility of applying confidence index and psychological safety boundary in vehicle motion control algorithm. The cost function of the model predictive controller in this section is composed of the tracking performance of the reference path and the subjective and objective evaluation model, as well as the potential field functions for obstacles and road boundaries. Consequently, the following is an expression of the evaluation model:

$$E\_confidence = f\_classifier \left( V\_dynamics \right) \qquad (1)$$

Where $E\_confidence$ is the subjective score of the driver, $V\_dynamics$ is a vector of each objective dynamic parameter listed in Table 1.

The total cost function of MPC can also include additional cost functions for obstacles and tracking reference paths. The following can be used to illustrate all cost functions of the control algorithm:

$$J_{all} = Q * J\_track + S * J\_obstacle + L * J\_road + R * J\_confidence \qquad (2)$$

In equation (2), the first term on the right of the equal sign denotes the deviation between the original reference trajectory and the predicted trajectory in the time domain; the second term represents the potential field function of obstacles; the third term represents the road potential field function; the fourth term indicates the subjective and objective evaluation model of confidence; Meanwhile, $Q$, $S$, $L$, and $R$ respectively depict the weights of each term in the total cost function.

The model predictive control has the benefit of making it simple to include various constraints in the control process. Model predictive control is chosen as the path tracking control algorithm in order to enable quick and precise tracking of the target path. Meanwhile, tracking control is mainly composed of longitudinal control and lateral control. The primary goal of this paper's study of lateral control is safe obstacle avoidance. A quadratic programming problem is created from the path tracking problem. Finally, the Active Set Method is used to obtain the control increment *(14)*. Corresponding sideslip angle, lateral acceleration, front wheel steering angle, and yaw rate are displayed in Figure 9. The red line in the top right figure represents the constraints.

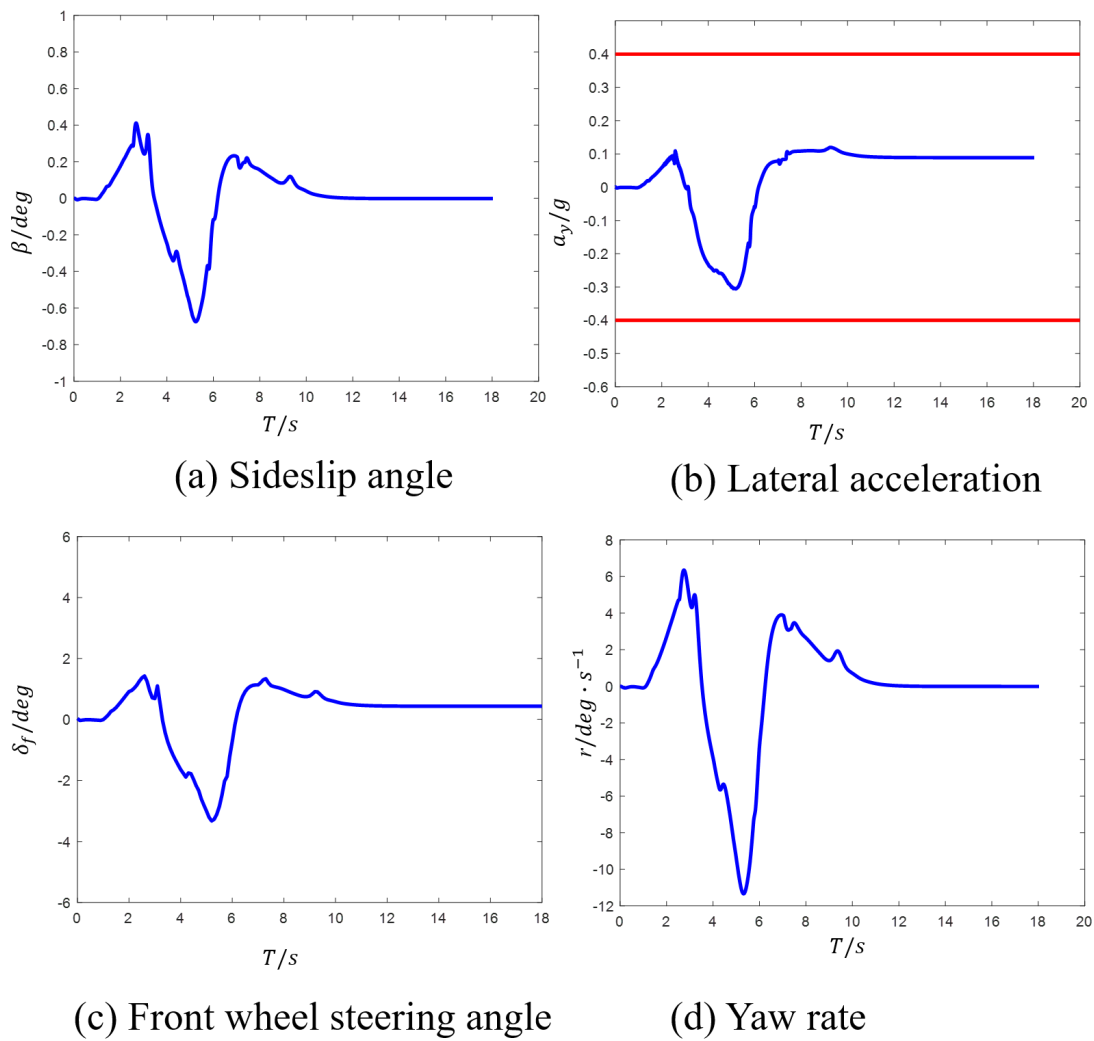

(a) Sideslip angle        (b) Lateral acceleration

(c) Front wheel steering angle        (d) Yaw rate

**Figure 9 Simulation results of vehicle dynamics parameters at 30km/h**

**RESULTS**

We conduct simulation research based on common obstacle avoidance settings to assess the efficacy of the evaluation model and control model based on human factors.

The static obstacle is in front of the controlled vehicle, and the vehicle completes obstacle avoidance through the left overtaking lane and moves back to the original lane to proceed. The control algorithm of autonomous vehicles is also consistent with the driving habits of real drivers when overtaking and avoiding obstacles. So, the autonomous vehicles can go back to the original lane to continue driving after overtaking in the passing lane.

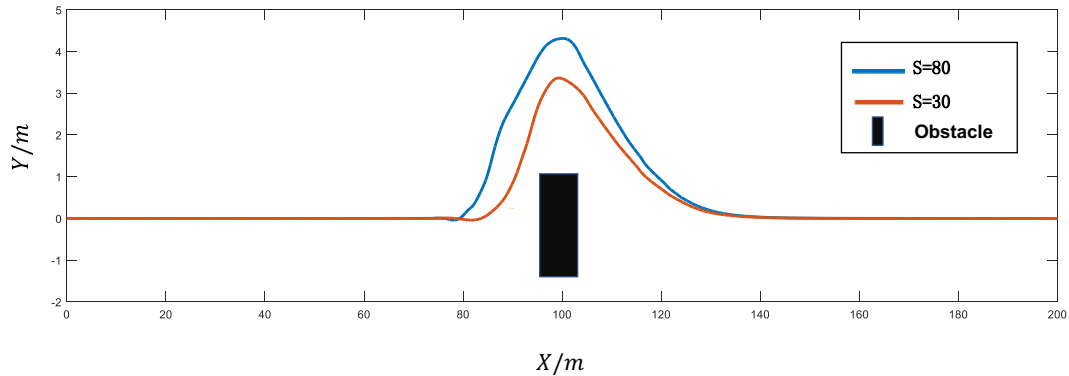

**Figure 10 Obstacle avoidance trajectories at different weights of obstacle potential field**

The influence of the obstacle on the vehicle motion control increases as the weight S of the obstacle potential field is increased from 30 to 80, as illustrated in Figure 10. Therefore, the steering operation is carried out at a distance from the obstacle. Meanwhile, the turning distance before obstacles is also different for drivers with different personalities.

The weight of the larger obstacle potential field (OPF) is more suitable for prudent drivers, while the weight of the smaller OPF is more suitable for reckless drivers. In conclusion, different weights of OPF can be applied to drivers with different driving styles.

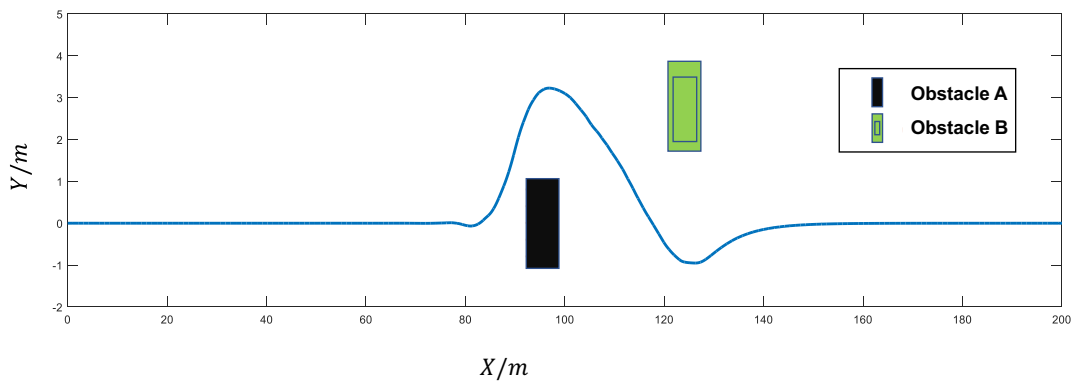

**Figure 11 Simulation results with double stationary obstacles**

This section chooses two static obstacles under cross-distributed working conditions, the obstacle potential field weight S = 20, the vehicle speed is chosen as 30 km/h, and the simulation results are displayed in Figure 11 to show whether the control algorithm is effective when the number and position of obstacles change. The outcomes demonstrate that the two cross-distributed obstacles can still be avoided by this controlled vehicle.

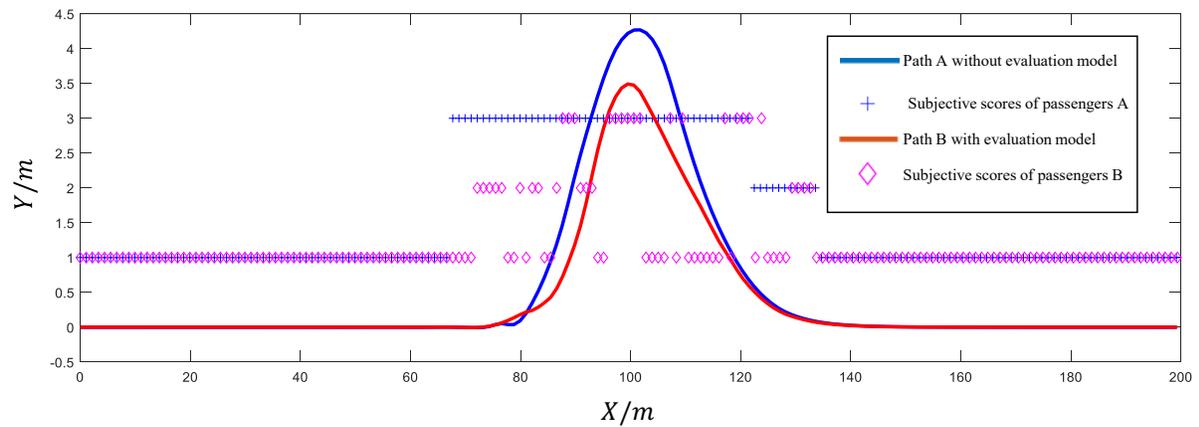

**Figure 12 Simulation results with/without evaluation model**

Figure 12 illustrates the two paths that emerged from the simulation. The evaluation model receives the dynamic parameters of each path as input, and outputs the subjective evaluation score. By comparing the simulation results of planner A without the evaluation model and planner B with the evaluation model, the results demonstrate that planner B can improve the driver's subjective feelings. First, the number of points with Y=1 of planner B is more than that of A, indicating that planner B has more points that represent the driver's psychological feeling as "good". Likewise, planner B has fewer "poor" points than planner A.

**TABLE 3 Subjective evaluation results with/without evaluation model**

| Planner | Subjective scores | | |
|---------|------|--------|------|
|         | Good | Normal | Poor |
| A       | 120  | 11     | 50   |
| B       | 149  | 16     | 16   |

According to Table 3, Planner B has more "good" points than A and fewer "poor" points than A. The results indicate that the addition of the evaluation model can improve the comfort of people in the obstacle avoidance process.

## CONCLUSIONS

Based on the psychological feelings of drivers and passengers when riding in autonomous vehicles, an evaluation model is proposed to characterize the human-machine trust feelings between drivers and autonomous driving systems. Four classifiers are used to establish an evaluation model, which represents the relationship between driver subjective feelings and vehicle dynamics parameters. According to the results, classifiers using Mahalanobis distance have the maximum accuracy, which is 68.18 percent. The established evaluation model is considered as one of the cost functions of MPC controller. The results show that the algorithm considering the evaluation model can effectively improve the driver's feelings.

**ACKNOWLEDGMENTS**

The authors would express their most sincere thanks to The Hong Kong University of Science Technology, Shanghai Jiao Tong University, and Pan-Asia Technical Automotive Center for their help.

**AUTHOR CONTRIBUTIONS**

The authors confirm contribution to the paper as follows: study conception and design: W. D., F. Y., Z. W., D. H.; data collection: W. D., F. Y., Z. W., D. H.; analysis and interpretation of results: W. D., F. Y., Z. W., D. H.; draft manuscript preparation: W. D., F. Y., Z. W., D. H. All authors reviewed the results and approved the final version of the manuscript.